# Semantics Meet Saliency: Exploring Domain Affinity and Models for Dual-Task Prediction

Md Amirul Islam[*,1]
amirul@cs.umanitoba.ca

Mahmoud Kalash[*,1]
kalashm@cs.umanitoba.ca

Neil D. B. Bruce[2]
bruce@ryerson.ca

[1] Department of Computer Science
University of Manitoba
Winnipeg, MB, Canada

[2] Department of Computer Science
Ryerson University
Toronto, ON, Canada

**Abstract**

Much research has examined models for prediction of semantic labels or instances including dense pixel-wise prediction. The problem of predicting salient objects or regions of an image has also been examined in a similar light. With that said, there is an apparent relationship between these two problem domains in that the composition of a scene and associated semantic categories is certain to play into what is deemed salient. In this paper, we explore the relationship between these two problem domains. This is carried out in constructing deep neural networks that perform both predictions together albeit with different configurations for flow of conceptual information related to each distinct problem. This is accompanied by a detailed analysis of object co-occurrences that shed light on dataset bias and semantic precedence specific to individual categories.

## 1 Introduction

While both semantic segmentation and saliency detection have been studied in detail, such study has mostly considered these problems in isolation. However, there is an intuitive conceptual relationship between these problem domains that has not been examined in detail to date. It is our view that there is much to be gained from careful study of the relationship between semantic segmentation and saliency both with respect to deep learning models, and also from a more general careful analysis of where these domains intersect. To this end, we aim to address a series of questions at the intersection of saliency and semantics and these questions, along with what we seek to understand are expressed in what follows:

**Does semantic segmentation help in predicting what is salient?** Intuitively one might expect that having a representation of image content at a semantic level has value for predicting what is salient. For example, a car might tend to be of greater interest than a tree on average. With that said, it is unclear to what extent such knowledge is of value or to what extent such concepts are instilled in a network trained to predict saliency alone.

[*]Both authors contributed equally to this work.

**Does saliency help with predicting fine-grained pixel-wise semantic labels?** Less clear than the role of semantics in predicting saliency is the converse case. It is not clear that knowing what may be of interest helps to define category. With that said, if a saliency model provides a more explicit representation of figure/ground segmentation, or contrast that helps to define object boundaries, this relationship may indeed prove to be useful.

**What sort of network configuration allows for the preceding objectives to be addressed most effectively?** In both of the aforementioned questions, there is the issue of how a network is configured, and how should information flow. For example, can the entire network rely on a common representation up to the highest layer and yield two separate predictions. It is also conceivable that different configurations may force training to emphasize one problem over another which might have a bearing on the extent to which one problem domain aids the other.

**Are there certain object categories that are always most salient?** Stepping outside of the realm of chasing the best possible score on benchmarks, there are some interesting general questions that reside at the intersection of saliency and semantics. For example, it is possible that if a particular semantic category is present, it is always the most salient. Observing whether this is the case requires careful analysis of such relationships. Moreover, there exists the possibility that this statement is true only because instances of a particular semantic category are typically from images where there are no competing figures of interest. These are questions that can be understood from the data.

**Are there any relationships between pairs of objects that show strong preference for one or the other?** Related to the prior question, is whether there exist co-occurrences that reveal any additional interesting patterns. For example, certain semantic categories may co-occur more often than others. Some may not co-occur at all. There may be a strong preference for one semantic category, or virtually no bias between a pair of semantic categories. In the latter case, it is conceivable that close examination of the data may reveal what tips the scales towards one class or the other. It is also possible that such examination brings to light certain biases in a dataset as opposed to general relationships.

These questions are addressed in the balance of paper through a series of experiments, and carefully chosen examples that shed light on where saliency and semantics meet.

## 2 Related Work

### 2.1 Semantic Segmentation

Convolutional neural network based approaches have shown tremendous success for the task of semantic segmentation due owing to the discriminative power of learned features. Earlier works on CNNs mostly relied on an end-to-end fully convolutional network [23] to accomplish pixel-wise prediction, despite the limitation of producing a low-resolution segmentation map. Directly upsampling the prediction map to the original size may degrade final prediction quality. To overcome this limitation, recent approaches employed the idea of dilated convolution [4, 30, 33], multi-scale processing [4, 7], encoder-decoder networks [3, 11, 12, 24], refinement based coarse-to-fine predictions [11, 12, 21, 25], Laplacian pyramids [6] and other similar strategies.

## 2.2 Saliency Detection

Early methods for salient object detection/segmentation mostly relied on hand-crafted features [14, 19, 29] including local or global contrast cues. Convolutional Neural Networks (CNNs) have raised the bar in performance for saliency detection in producing more capable feature detectors to determine salient regions. Some methods exploited superpixel or bounding box proposals [10, 15, 16, 17, 18, 34] to predict salient regions. Later efforts take advantage of CNNs also using strategies that include combining multi-scale features [31], encoder-decoder pipelines [22, 27, 32], and refinement strategies[2, 22, 31].

Architecturally, our work is closest to the strategy proposed in CSM [1] which predicts salient regions by applying semantic labels, while in this work we aim to predict semantic and salient regions simultaneously by jointly training the network for both tasks. Recent work [26] also attempted to predict salient regions by incorporating image-level groundtruth in a weakly-supervised manner. In contrast to the above-described approaches, we achieve semantic segmentation and saliency detection from a common source network and establish the concept of semantics driven saliency by comprehensive experiments.

# 3 Semantic Driven Saliency

In this section, we propose an end-to-end framework for solving the problem of semantic segmentation and saliency detection simultaneously.

## 3.1 Key Observations

Previous works [2, 4, 9, 12, 21, 22, 23, 28, 31, 32, 33] leverage CNNs to predict semantic label for each pixel or alternatively, to detect the salient objects. A few existing approaches [1, 26] take advantage of pixel-wise semantic labels/image-level labels to predict salient objects in an unsupervised manner. In our approach, we aim to use both sets of labels (semantic and saliency) to improve prediction for both of the tasks. What we have common with current literature is that initially we also use a fully convolutional network as a backbone to extract a rich feature representation for both tasks. This canonical design seems to work well for semantic segmentation or saliency detection separately, but it is unclear how such a model might behave when applied to both tasks simultaneously. The fundamental reason behind that is the fact that feature representation at a high level may not be powerful enough to deal with both tasks and, e.g., might be biased more towards representing semantic categories than saliency. Taking into account the nature of semantic segmentation and saliency detection, a natural way to build our network to extract more advanced feature representations is to add task-specific branches assuming the same backbone network. In this light, four different variants are considered.

## 3.2 Overview of the Proposed Framework

An illustration of our proposed pipeline is shown in Fig. 1. The CNNs serves as a global feature extractor that transforms the input image to a rich feature representation, while the task-specific stages aim to extract powerful features that are capable of predicting labels precisely. In the following subsections, we show how to build a generalized structure that can solve both tasks simultaneously.

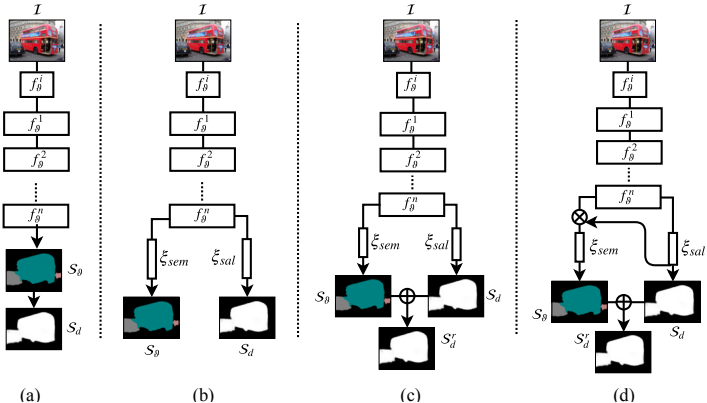

Figure 1: Illustration of different network architecture variants. These are chosen to control task-emphasis in considering performance on either of the target tasks.

## 3.3 Semantics Driven Saliency Network

Recent feed-forward models applied to semantic segmentation or saliency detection employ a fully convolutional network comprised of repeated convolutional stages and spatial pooling. The network is capable of achieving highly detailed semantic features at the deepest stage of the network. Since pixel-wise labeling tasks require pixel-precise information, following prior work [2, 12, 21, 33], we chose the dilated ResNet101 [4] as our backbone network considering its very capable performance. The network produces a $\frac{1}{8}$ sized output feature map compared to the original input. More specifically, given an input image $I \in \mathbb{R}^{h \times w \times c}$, the backbone network produces a feature map of size $\lfloor \frac{h}{8}, \frac{w}{8} \rfloor$. A natural way to construct a general architecture is to employ the same network (backbone) for both the tasks (Fig. 1 (a)). In this case, the network predicts semantic labels ($S_\vartheta$) first and then subsequently produces saliency predictions ($S_d$) directly from the semantic labels by applying a 3 × 3 convolution. As mentioned earlier, due to lack of discriminative features at high-levels the direct dependency between semantic labels in saliency may result in a noisy prediction. In considering a structure where this dependency is removed, we create two separate prediction branches that take the encoded feature map as input. However, in this configuration, the network fails to predict accurate labeling for both tasks as both the prediction layers use the same backbone network. Considering this limitation, we propose to include task-specific branches at the top of the backbone network. So, we produce a network configuration (Fig. 1 (b)) that applies a CNN common to both the tasks. We then add two different branches ($\xi_{sem}$ and $\xi_{sal}$) specific to each task followed by a prediction layer at the end of each branch. Note that the two branches have similar layer configurations but without sharing the weights, allowing the branches to learn task-specific features. To further expand the interaction between semantic and saliency predictions, we propose another configuration (Fig. 1 (c)) that is similar to (Fig. 1 (b)) except we concatenate the semantic and saliency predictions to produce a refined saliency map ($S_d^r$). The reasoning behind this is that the initial prediction layer allows the network branch specialized for saliency to benefit from reasoning about semantics without explicit reliance on the same features. Finally, to facilitate semantic learning driven by saliency, we use the feature-level gating concept introduced in [12] as shown in (Fig. 1 (d)). More specifically, we concatenate the output features of the saliency branch to the input feature map of the semantic branch. Then we attach a 1×1 convolution layer with channels similar to the expected input channel of the semantic branch.

## 3.4 Training the Network

We start with a set of training images, denoted as $\mathbb{X} = \{x_i, \text{where } i = 1, 2, ..., N\}$, a set of semantic ground-truth maps, and a set of saliency ground-truth maps. In more specific terms, let $I \in \mathbb{R}^{h \times w \times 3}$ be a training image with ground-truth semantic map $\mathcal{S}_m \in \mathbb{R}^{h \times w}$ and saliency map $\mathcal{G}_m \in \mathbb{R}^{h \times w}$. To apply supervision on the predicted maps, we first downsample $\mathcal{S}_m$ and $\mathcal{G}_m$ to the size of $S_\vartheta$. Then we define pixel-wise cross entropy loss $\Delta_{\xi_{sem}}$ and $\Delta_{\xi_{sal}}$ to measure the difference between $(S_\vartheta, \mathcal{S}_m)$ and $(S_d, \mathcal{G}_m)$ respectively. We can summarize these operations as:

$$\Delta_{\xi_{sem}}(W) = \alpha(S_\vartheta, \mathcal{S}_m), \quad \Delta_{\xi_{sal}}(W) = \alpha(S_d, \mathcal{G}_m) \tag{1}$$

where $\alpha$ denotes the cross-entropy loss. We also apply supervision on the refined saliency map on version II and III. So, the final loss function of the network combining semantic and saliency losses can be written as follows:

$$L_{final}(W) = \Delta_{\xi_{sem}}(W) + \Delta_{\xi_{sal}}(W) + \Delta_{\xi_{sal}^r}(W) \tag{2}$$

For semantic and saliency inference, we feed an input image of original resolution to the trained network and obtain the semantic and saliency predictions.

# 4 Experimental Results

In this section, we report experimental results for semantic and saliency detection tasks on the PASCAL-S dataset [20].

## 4.1 Setup

**Implementation Details:** We use the publicly available Caffe framework [13] to implement our proposed network. We build our network upon the publicly available DeepLab network [4] which follows a dilated structure based on ResNet-101 [8].
**Dataset:** The PASCAL-S dataset is one of the most commonly used salient object detection datasets and is unique in having multiple explicitly tagged salient regions. Unlike other saliency detection datasets where the most salient object dominates the image, PASCAL-S provides ground-truth based on multi-observer agreement across several objects. It contains a total of 850 images derived from the PASCAL VOC 2011 validation set [5]. Following [2], we randomly split the PASCAL-S dataset into two sets (425 for training and 425 for testing). Note that PASCAL-S dataset does not provide semantic annotation consistent to PASCAL VOC 2011. We therefore collect category-wise semantic annotations for PASCAL-S images from the Pascal VOC 2011 dataset. Similar to existing approaches for salient object segmentation or detection, the ground-truth saliency map is thresholded and treated as a binary representation.
**Evaluation Metrics:** We use three standard metrics to measure saliency detection performance including F-measure (max along the curve), Area under ROC curve (AUC), and mean absolute error (MAE). For the semantic segmentation task, we report pixel accuracy, category-wise mean accuracy, and mean IoU.

## 4.2 Experimental results on PASCAL-S dataset

Table 1 shows the quantitative results for all versions of our model, and comparison with the baseline models. It is clear that the proposed approaches (SDS-v2 & SDS-v3) achieve better performance for both semantic segmentation and salient object detection tasks. From Table 1, we can see that SDS-v2 improves the mean IoU by a considerable margin, whereas SDS-v3 outperforms the saliency baseline that is only trained on saliency annotation by a reasonable margin in terms of $F_m$, AUC, and MAE. As mentioned earlier, training a straightforward CNN (SDS-v1) to predict semantic labels and salient regions sequentially is insufficient to achieve better performance than baseline (*0.658* vs *0.656* mIoU).

| Methods | Semantic | | | Saliency | | |
|---|---|---|---|---|---|---|
| | Pixel Acc. | Mean Acc. | mIoU | $F_m$ | AUC | MAE |
| Baseline-Sem [4] | 0.940 | 0.733 | 0.658 | - | - | - |
| Baseline-Sal [4] | - | - | - | 0.743 | 0.94 | **0.079** |
| SDS-v1 | 0.939 | 0.731 | 0.656 | 0.74 | 0.948 | 0.081 |
| SDS-v2 | **0.942** | **0.742** | **0.666** | 0.738 | **0.951** | 0.083 |
| SDS-v3 | 0.940 | 0.742 | 0.662 | **0.744** | 0.95 | **0.079** |
| SDS-v4 | 0.939 | 0.741 | 0.661 | 0.738 | 0.95 | 0.082 |

Table 1: Results of semantic segmentation and saliency detection on PASCAL-S dataset by joint training. To obtain a fair assessment of performance gain, we compare with two approaches that use a similar base network but are trained for independent tasks.

Fig. 2 depicts a qualitative comparison on the PASCAL-S dataset. We can see that the proposed approach is more precise in predicting semantic and salient regions.

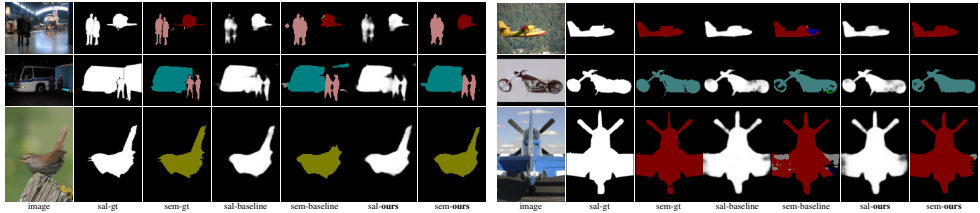

Figure 2: Qualitative comparison of results on the PASCAL-S dataset. Our approach produces more accurate saliency and segmentation maps compared to baseline.

## 4.3 Semantic and Saliency Analysis

In this section, we explore semantic-saliency co-occurrences and how the representation of saliency may drive more accurate semantic maps and vice versa. Since category-wise semantic annotation (as well as saliency maps) in PASCAL-S are not consistent with annotations provided by Pascal VOC 2011 in terms of the mask corresponding to a given instance, we can not directly use the two sets of annotations to generate a rank order of semantic categories. Note that for this analysis, we examine saliency in considering rank rather than as a binary quantity [2]. To obtain a relative rank order, we combine the saliency map with the semantic annotation from PASCAL VOC 2011 to assign a rank order to semantic categories. Directly combining these two sources of annotations to achieve the desired objective is a more significant challenge than one might expect. There are many instances which are labeled in the saliency map in PASCAL-S but are not labeled as a semantic category in PASCAL VOC 2011. Other challenges include categories (multiple instances of person) ranked differently

**Algorithm 1** Relative Rank of Semantic Categories
  **function** SEMANTICRANK($\mathcal{I}, \mathcal{F}$)
    ▷ class-wise semantic maps $\mathcal{I}$, saliency maps $\mathcal{F}$
    **for** each semantic category $\mathcal{I}_i \in \mathcal{I}$ **do**
      overlap, $\vartheta_i = \mathcal{F}_i \times \mathcal{I}_i$
      valid overlap, $\vartheta'_i = compare(\vartheta_i, \mathcal{F}_i)$
      $\mathbb{R}_m = \max(\vartheta'_i)$,   $class\text{Rank}(I_i) = \mathbb{R}_m$
    **end for**
  **end function**

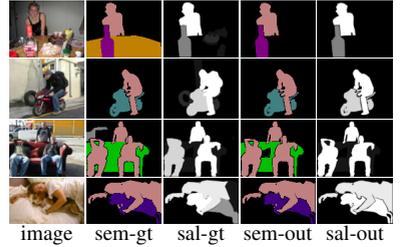

Input and output samples

image    sem-gt    sal-gt    sem-out    sal-out

across the dataset or segments of one semantic category assigned to a different one in the saliency map and vice versa.

To overcome these limitations, we have arrived at a data refinement pipeline through experimentation that is shown to produce faithful rankings for semantic categories. The method (see Algorithm 1) that address this challenge is briefly described in what follows. We first obtain the semantic maps $\mathcal{I}$ and saliency maps $\mathcal{F}$. For each semantic category, we calculate the overlap $\vartheta$ by element-wise multiplication with the saliency map $\mathcal{I}$. Once we obtain the overlap for each category, we apply an inclusion/rejection criterion to get a cleaner overlap mask. We then *compare* the original category mask size with the overlapping one to remove/keep only the valid instances (see the first row of the algorithm samples).

| categories | aero | bike | bird | boat | bottle | bus | car | cat | chair | cow | table | dog | horse | mbike | person | plant | sheep | sofa | train | tv |
|---|---|---|---|---|---|---|---|---|---|---|---|---|---|---|---|---|---|---|---|---|
| overall | 57 | 53 | 73 | 11 | 60 | 44 | 80 | 85 | 41 | 45 | 32 | 81 | 50 | 52 | 241 | 48 | 35 | 25 | 51 | 42 |
| salient | 56 | 49 | 73 | 8 | 54 | 43 | 66 | 85 | 23 | 45 | 8 | 81 | 50 | 52 | 234 | 33 | 35 | 13 | 50 | 36 |
| salient alone | 50 | 22 | 68 | 0 | 28 | 23 | 34 | 69 | 0 | 39 | 0 | 58 | 37 | 21 | 63 | 15 | 32 | 1 | 45 | 24 |
| Distrib | 0.98 | 0.92 | 1.00 | 0.73 | 0.90 | 0.98 | 0.83 | 1.00 | 0.56 | 1.00 | 0.25 | 1.00 | 1.00 | 1.00 | 0.97 | 0.69 | 1.00 | 0.52 | 0.98 | 0.86 |
| **Rank-1** | 54 | 32 | 72 | 4 | 34 | 39 | 41 | 83 | 4 | 44 | 2 | 77 | 47 | 38 | 183 | 17 | 34 | 4 | 49 | 31 |
| **Rank-1(%)** | 0.95 | 0.6 | 0.99 | 0.36 | 0.57 | 0.89 | 0.51 | 0.98 | 0.1 | 0.98 | 0.06 | 0.95 | 0.94 | 0.73 | 0.76 | 0.35 | 0.97 | 0.16 | 0.96 | 0.74 |
| **Rank-2** | 2 | 15 | 1 | 3 | 19 | 4 | 18 | 1 | 16 | 1 | 5 | 4 | 3 | 14 | 50 | 15 | 1 | 6 | 1 | 5 |
| **Rank-2(%)** | 0.04 | 0.28 | 0.01 | 0.27 | 0.32 | 0.09 | 0.23 | 0.01 | 0.39 | 0.02 | 0.16 | 0.05 | 0.06 | 0.27 | 0.21 | 0.31 | 0.03 | 0.24 | 0.02 | 0.12 |
| **Rank-3** | 0 | 2 | 0 | 1 | 0 | 0 | 7 | 1 | 3 | 0 | 1 | 0 | 0 | 0 | 1 | 1 | 0 | 3 | 0 | 0 |
| **Rank-3(%)** | 0 | 0.04 | 0 | 0.09 | 0 | 0 | 0.09 | 0.01 | 0.07 | 0 | 0.03 | 0 | 0 | 0 | 0 | 0.02 | 0 | 0.12 | 0 | 0 |

Table 2: Distribution of semantic categories appearing overall, and as a salient instance across the dataset. Last three rows denote the total number of times a category appeared as rank-1 (most salient), rank-2, and rank-3 respectively.

### 4.3.1 Distribution of Semantic Categories Across Dataset

Initially, we obtain respectively: (i) the distribution of semantic categories across the dataset that includes overall frequency of appearance disregarding saliency, (ii) how often a category appeared as salient (rank does not matter), and (iii) appearance as a salient object with no other objects in the image. Table 2 shows these distributions. It is an evident from the table that *person* is the most salient category across the dataset followed by *cat* and *car*. Additionally, regarding quantitative rank order person dominates the others. Interestingly, even though the saliency occupancy of other categories (e.g. *train*, *cat*, *cow*) is significantly less then *person*, the probability of appearing as top-ranked salient object given that it appears as salient is relatively higher (0.96 vs 0.76) than *person*. Also, animal categories (e.g., *cat, dog, sheep, horse, cow, bird*) tend to be most salient categories whenever they appear in the image. Fig. 3 shows the statistics of rank order for each semantic category.

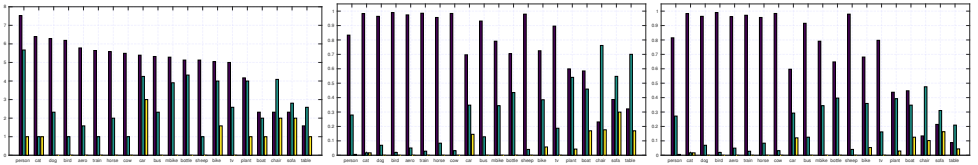

Figure 3: Statistics for semantic categories. Colors correspond to objects ranked as most salient (Blue), second most salient (Green) or third most salient (Yellow). Left: Total count across all images. Middle: Total count relative to number of images that include a salient instance of the object. Right: Total count relative to number of images that include an instance of the object. For contrast, a log-scale is used.

### 4.3.2 Co-occurrence of Semantic Categories

We also consider co-occurrences among semantic categories to examine guiding principles in how salient objects are chosen and how these relate to semantic categories. We report the number of times two categories co-occur as salient in Table 3(a). We also report the chances of each category being most salient when co-occurring with all others.

| * | aero | bike | bird | boat | bottle | bus | car | cat | chair | cow | table | dog | horse | mbike | person | plant | sheep | sofa | train | tv |
|---|---|---|---|---|---|---|---|---|---|---|---|---|---|---|---|---|---|---|---|---|
| aero | x | 0 | 0 | 0 | 0 | 1 | 1 | 0 | 0 | 0 | 0 | 0 | 0 | 0 | 4 | 0 | 0 | 0 | 0 | 0 |
| bike | 0 | x | 0 | 0 | 4 | 1 | 6 | 1 | 0 | 0 | 0 | 0 | 0 | 0 | 22 | 1 | 0 | 0 | 0 | 1 |
| bird | 0 | 0 | x | 0 | 0 | 1 | 0 | 0 | 0 | 0 | 0 | 0 | 0 | 1 | 3 | 1 | 0 | 0 | 0 | 0 |
| boat | 0 | 0 | 0 | x | 0 | 0 | 0 | 0 | 0 | 0 | 0 | 0 | 0 | 1 | 8 | 0 | 0 | 0 | 0 | 0 |
| bottle | 0 | 4 | 0 | 0 | x | 1 | 2 | 2 | 1 | 0 | 0 | 0 | 0 | 0 | 21 | 0 | 0 | 1 | 0 | 1 |
| bus | 1 | 1 | 1 | 0 | 1 | x | 5 | 0 | 0 | 0 | 0 | 0 | 0 | 0 | 14 | 0 | 0 | 0 | 0 | 0 |
| car | 1 | 6 | 0 | 0 | 2 | 5 | x | 0 | 0 | 0 | 0 | 2 | 0 | 6 | 20 | 2 | 0 | 0 | 0 | 0 |
| cat | 0 | 1 | 0 | 0 | 2 | 0 | 0 | x | 2 | 1 | 1 | 0 | 0 | 0 | 7 | 3 | 0 | 3 | 0 | 2 |
| chair | 0 | 0 | 0 | 0 | 1 | 0 | 0 | 2 | x | 0 | 5 | 2 | 0 | 0 | 9 | 6 | 0 | 2 | 0 | 2 |
| cow | 0 | 0 | 0 | 0 | 0 | 0 | 0 | 1 | 0 | x | 0 | 0 | 0 | 0 | 6 | 0 | 0 | 0 | 0 | 0 |
| table | 0 | 0 | 0 | 0 | 0 | 0 | 0 | 1 | 5 | 0 | x | 1 | 0 | 0 | 2 | 1 | 0 | 1 | 0 | 0 |
| dog | 0 | 0 | 0 | 0 | 0 | 0 | 2 | 0 | 2 | 0 | 1 | x | 0 | 0 | 18 | 1 | 1 | 0 | 0 | 0 |
| horse | 0 | 0 | 0 | 0 | 0 | 0 | 0 | 0 | 0 | 0 | 0 | 0 | x | 0 | 13 | 0 | 0 | 0 | 0 | 0 |
| mbike | 0 | 0 | 0 | 1 | 0 | 0 | 6 | 0 | 0 | 0 | 0 | 0 | 0 | x | 28 | 0 | 1 | 0 | 0 | 0 |
| person | 4 | 22 | 3 | 8 | 21 | 14 | 20 | 7 | 9 | 6 | 2 | 18 | 13 | 28 | x | 5 | 1 | 5 | 5 | 5 |
| plant | 0 | 1 | 1 | 0 | 0 | 0 | 2 | 3 | 6 | 0 | 1 | 1 | 0 | 0 | 5 | x | 0 | 4 | 0 | 3 |
| sheep | 0 | 0 | 0 | 0 | 0 | 0 | 0 | 0 | 0 | 1 | 0 | 1 | 0 | 1 | 1 | 0 | 35 | 0 | 0 | 0 |
| sofa | 0 | 0 | 0 | 1 | 0 | 0 | 3 | 2 | 0 | 1 | 0 | 0 | 0 | 0 | 5 | 4 | 0 | x | 0 | 3 |
| train | 0 | 0 | 0 | 0 | 0 | 0 | 0 | 0 | 0 | 0 | 0 | 0 | 0 | 0 | 5 | 0 | 0 | 0 | x | 0 |
| tv | 0 | 1 | 0 | 0 | 1 | 0 | 0 | 2 | 2 | 0 | 0 | 0 | 0 | 0 | 5 | 3 | 0 | 3 | 0 | x |

(a)

| * | aero | bike | bird | boat | bottle | bus | car | cat | chair | cow | table | dog | horse | mbike | person | plant | sheep | sofa | train | tv |
|---|---|---|---|---|---|---|---|---|---|---|---|---|---|---|---|---|---|---|---|---|
| aero | 0 | 0 | 0 | 0 | 0 | 1 | 1.0 | 0 | 0 | 0 | 0 | 0 | 0 | 0 | 0.3 | 0 | 0 | 0 | 0 | 0 |
| bike | 0 | 0 | 0 | 0 | 0.5 | 0 | 0.8 | 0 | 0 | 0 | 0 | 0 | 0 | 0 | 0.1 | 1 | 0 | 0 | 0 | 0 |
| bird | 0 | 0 | 0 | 0 | 0 | 0 | 0 | 0 | 0 | 0 | 0 | 0 | 0 | 0 | 0.7 | 1.0 | 0 | 0 | 0 | 0 |
| boat | 0 | 0 | 0 | 0 | 0 | 0 | 0 | 0 | 0 | 0 | 0 | 0 | 0 | 0 | 0.3 | 0 | 0 | 0 | 0 | 0 |
| bottle | 0 | 0 | 0 | 0 | 0 | 0 | 0.5 | 0 | 1.0 | 0 | 0 | 0 | 0 | 0 | 0.1 | 0 | 0 | 1.0 | 0 | 0 |
| bus | 0 | 1.0 | 0 | 0 | 0 | 0 | 0.8 | 0 | 0 | 0 | 0 | 0 | 0 | 0 | 0.4 | 0 | 0 | 0 | 0 | 0 |
| car | 0 | 0.2 | 0 | 0 | 0 | 0 | 0 | 0 | 0 | 0 | 0 | 0.2 | 0 | 0 | 0.3 | 0 | 0 | 0 | 0 | 0 |
| cat | 0 | 1.0 | 0 | 0 | 0.5 | 0 | 0 | 0 | 1.0 | 0 | 1.0 | 0 | 0 | 0 | 0.4 | 1.0 | 0 | 1.0 | 0 | 1.0 |
| chair | 0 | 0 | 0 | 0 | 0 | 0 | 0 | 0 | 0 | 0 | 0 | 0 | 0 | 0 | 0.4 | 0.5 | 0 | 0 | 0 | 0 |
| cow | 0 | 0 | 0 | 0 | 0 | 0 | 0 | 1.0 | 0 | 0 | 0 | 0 | 0 | 0 | 0.7 | 0 | 0 | 0 | 0 | 0 |
| table | 0 | 0 | 0 | 0 | 0 | 0 | 0 | 0.4 | 0 | 0 | 0 | 0 | 0 | 0 | 0 | 0 | 0 | 0 | 0 | 0 |
| dog | 0 | 0 | 0 | 0 | 0 | 0 | 1.0 | 0 | 1.0 | 0 | 1.0 | 0 | 0 | 0 | 0.6 | 1.0 | 0 | 0 | 0 | 0 |
| horse | 0 | 0 | 0 | 0 | 0 | 0 | 0 | 0 | 0 | 0 | 0 | 0 | 0 | 0 | 0.3 | 0 | 0 | 0 | 0 | 0 |
| mbike | 0 | 0 | 0 | 1.0 | 0 | 0 | 1.0 | 0 | 0 | 0 | 0 | 0 | 0 | 0 | 0.1 | 0 | 1.0 | 0 | 0 | 0 |
| person | 0.5 | 0.7 | 0.3 | 0.5 | 0.9 | 0.1 | 0.7 | 0.3 | 1 | 0.2 | 1.0 | 0.2 | 0.2 | 0.5 | 0 | 1.0 | 0 | 1.0 | 0.2 | 0.6 |
| plant | 0 | 0 | 0 | 0 | 0 | 0 | 0 | 1.0 | 0.3 | 0 | 1.0 | 0 | 0 | 0 | 0 | 0 | 0 | 0.3 | 0 | 0 |
| sheep | 0 | 0 | 0 | 0 | 0 | 0 | 0 | 0 | 0 | 0 | 0 | 0 | 0 | 1.0 | 0 | 0 | 0 | 0 | 0 | 0 |
| sofa | 0 | 0 | 0 | 0 | 0 | 0 | 1.0 | 0 | 1.0 | 0 | 0 | 0 | 0 | 0 | 0 | 0.5 | 0 | 0 | 0 | 0.3 |
| train | 0 | 0 | 0 | 0 | 0 | 0 | 0 | 0 | 0 | 0 | 0 | 0 | 0 | 0 | 0.8 | 0 | 0 | 0 | 0 | 0 |
| tv | 0 | 1.0 | 0 | 0 | 1.0 | 0 | 0 | 0 | 0 | 0 | 0 | 0 | 0 | 0 | 0.4 | 0.7 | 0 | 0.7 | 0 | 0 |

(b)

Table 3: (a) Co-occurrence matrix for semantic categories appearing as salient. (b) Co-occurrence matrix of semantic categories appearing as salient. Numbers indicate the probability that the row category is ranked higher than the column category.

**A case study:** We perform a deeper analysis with the most frequent salient category (*person*) and its top 7 co-occurring salient categories represented in Table 4. We can see that motorbike has the most co-occurrences (28 times) with person compared to others. We also report the chances of the person being most salient when co-occurring with other objects. For instance, in a *person* vs. *motorbike* co-occurrence there is 50% chance person will have higher rank than motorbike and only 14% chance motorbike will have higher rank. The graph representation in right of Table 4 shows the co-occurrence of person with the top 7 categories where the thickness of the node and edges denote how often a co-occurrence is observed and how often person is favored over the others respectively. An interesting observation from Table 4 is that the person tends to be ranked lower when it co-exists with *dog, bus, horse*. In order to explain this, we do a further investigation of those cases. Fig. 4 shows some samples where this happens. It is evident from the images that the person rank might be affected by other factors in the image. More specifically, the person tends to be ranked lower due to: central bias (first two images in the top row), person is in the background, too small, or not completely present in the image. Action in the image is more likely to gain attention than a specific category, and people lying down seem to draw less interest.

| * | mbike | bike | bottle | car | dog | bus | horse |
|---|---|---|---|---|---|---|---|
| Salient (with **person**) | **28** | 22 | 21 | 20 | 18 | 14 | 13 |
| % **person** | 0.50 0.14 | 0.68 0.14 | 0.86 0.10 | 0.65 0.25 | 0.17 0.56 | 0.14 0.36 | 0.13 0.23 |

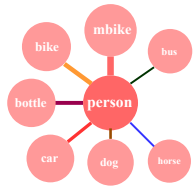

Table 4: Co-occurrence of *person* with other frequent semantic categories. Red color represents the probability that the row category (i.e person in all cases) is ranked higher than the column category whereas the blue color represents the probability that the column category is ranked higher than the row.

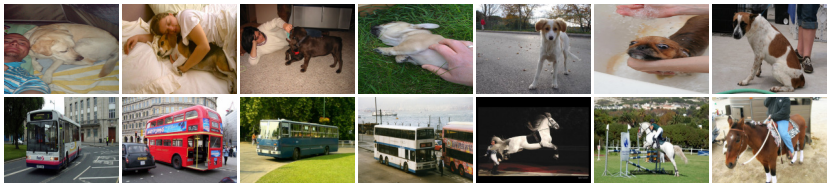

Figure 4: Samples of images where person is ranked lower than the co-occurring category.

### 4.3.3 Categorial Improvement with Saliency Influence

To justify the value of saliency driven semantics, we report category-wise mean IoU (*baseline* vs *ours*) in Table 5. It is evident from Table 5 that IoU for most of the categories that appeared as salient improved significantly. As mentioned earlier, *person* tends to be the most salient category (as shown experimentally) resulting in a noticeable improvement on IoU (*0.84 vs. 0.86*). Similarly, our proposed approach achieves higher IoU on other categories that appear as salient objects compared to baseline.

| method | aero | bike | bird | boat | bottle | bus | car | cat | chair | cow | table | dog | horse | mbike | person | plant | sheep | sofa | train | tv | mIoU |
|---|---|---|---|---|---|---|---|---|---|---|---|---|---|---|---|---|---|---|---|---|---|
| Semantic [4] | 0.88 | 0.6 | 0.84 | 0.15 | **0.69** | 0.89 | 0.77 | 0.84 | 0.18 | 0.56 | **0.44** | 0.75 | **0.67** | 0.77 | **0.84** | 0.6 | 0.73 | 0.24 | 0.78 | 0.69 | 0.66 |
| SDS-v3 | 0.87 | 0.56 | 0.85 | 0.14 | **0.72** | 0.88 | 0.77 | 0.85 | 0.19 | 0.59 | **0.49** | 0.75 | **0.71** | 0.77 | **0.86** | 0.61 | 0.73 | 0.24 | 0.76 | 0.69 | **0.67** |

Table 5: Quantitative results in terms of mean IoU on PASCAL-S. Note that improvements are especially pronounced for categories that appear most frequently as highly salient.

## 5 Conclusion

In this paper, we have explored the relationship between semantics and saliency. In doing so, we have presented four variants of an end-to-end deep learning framework for simultaneously predicting semantic labels and salient regions, each aimed at testing domain dependencies. We demonstrate that semantics and saliency can be combined to mutual benefit of performance for both tasks. Comprehensive experiments demonstrate effectiveness, and we also provide an in depth analysis of relationships among semantic categories, and explanations for cross-semantic relationships that are observed.

**Acknowledgements:** The authors gratefully acknowledge financial support from the NSERC Discovery Grants program and Manitoba Graduate Scholarship. We also thank the NVIDIA Corporation for providing GPUs through their academic program.